\def\year{2021}\relax
\documentclass[letterpaper]{article} 
\usepackage{aaai21}  
\usepackage{times}  
\usepackage{helvet} 
\usepackage{courier}  
\usepackage[hyphens]{url}  
\usepackage{graphicx} 
\usepackage{natbib}  
\usepackage{caption} 
\usepackage{amsfonts}       
\usepackage{nicefrac}       
\usepackage{microtype}      

\usepackage{epsfig}
\usepackage{amsmath}
\usepackage{amssymb}
\usepackage{epstopdf}
\usepackage{color}
\usepackage{array}
\usepackage{comment}
\usepackage{algorithm}
\usepackage{algorithmicx}
\usepackage{cases}
\usepackage{algpseudocode}
\usepackage{multirow}
\usepackage{textcomp}
\usepackage{subfigure}
\usepackage{booktabs}
\usepackage[export]{adjustbox}
\usepackage{bbm}
\newcommand{\eg}{\textit{e.g.}}

\newcommand{\etal}{\textit{et al}}

\title{A Simple Yet Effective Method for Video Temporal Grounding with Cross-Modality Attention}
\author{
    Binjie Zhang\textsuperscript{\rm 1}\thanks{This work was done when Binjie's intern at Tencent.},
    Yu Li\textsuperscript{2},
    Chun Yuan\textsuperscript{1},
    Dejing Xu\textsuperscript{2},
    Pin Jiang\textsuperscript{3},
    Ying Shan\textsuperscript{2}
    \\
}
\affiliations{
    \textsuperscript{\rm 1}Tsinghua Shenzhen International Graduate School\\
    \textsuperscript{2}Applied Research Center (ARC), Tencent PCG\\
    \textsuperscript{3}Tianjin University
    zbj19@mails.tsinghua.edu.cn, yuanc@sz.tsinghua.edu.cn, jpin@tju.edu.cn, \{ianyli,djxu,yingshan\} @tencent.com
}

\begin{document}
\maketitle

\begin{abstract}
The task of language-guided video temporal grounding is to localize the particular video clip corresponding to a query sentence in an untrimmed video. Though progress has been made continuously in this field, some issues still need to be resolved. First, most of the existing methods rely on the combination of multiple complicated modules to solve the task. Second, due to the semantic gaps between the two different modalities, aligning the information at different granularities (local and global) between the video and the language is significant, which is less addressed. Last, previous works do not consider the inevitable annotation bias due to the ambiguities of action boundaries. To address these limitations, we propose a simple two-branch Cross-Modality Attention (CMA) module with intuitive structure design, which alternatively modulates two modalities for better matching the information both locally and globally. In addition, we introduce a new task-specific regression loss function, which improves the temporal grounding accuracy by alleviating the impact of annotation bias. We conduct extensive experiments to validate our method, and the results show that just with this simple model, it  can outperform the state of the arts on both Charades-STA and ActivityNet Captions datasets.
\end{abstract}

\section{Introduction}
    \noindent Vision and language are two of the most important representations of information. With the development of computer vision and natural language processing, multi-modality tasks have also drawn increasing attention, such as video understanding ~\cite{ma2019ts, zolfaghari2018eco}, video caption ~\cite{VC:zhou2018end, VC:iashin2020better}, text-to-video retrieval ~\cite{VR:mithun2018learning, VR:chen2020fine} and video question answer ~\cite{VQA:cadene2019murel,VQA:gao2019multi}. The core problem among these tasks involves the multi-modality fusion and interaction, which still needs to be resolved due to the semantic gaps.

\begin{figure} [t]
    \centering
    \includegraphics[width=\linewidth]{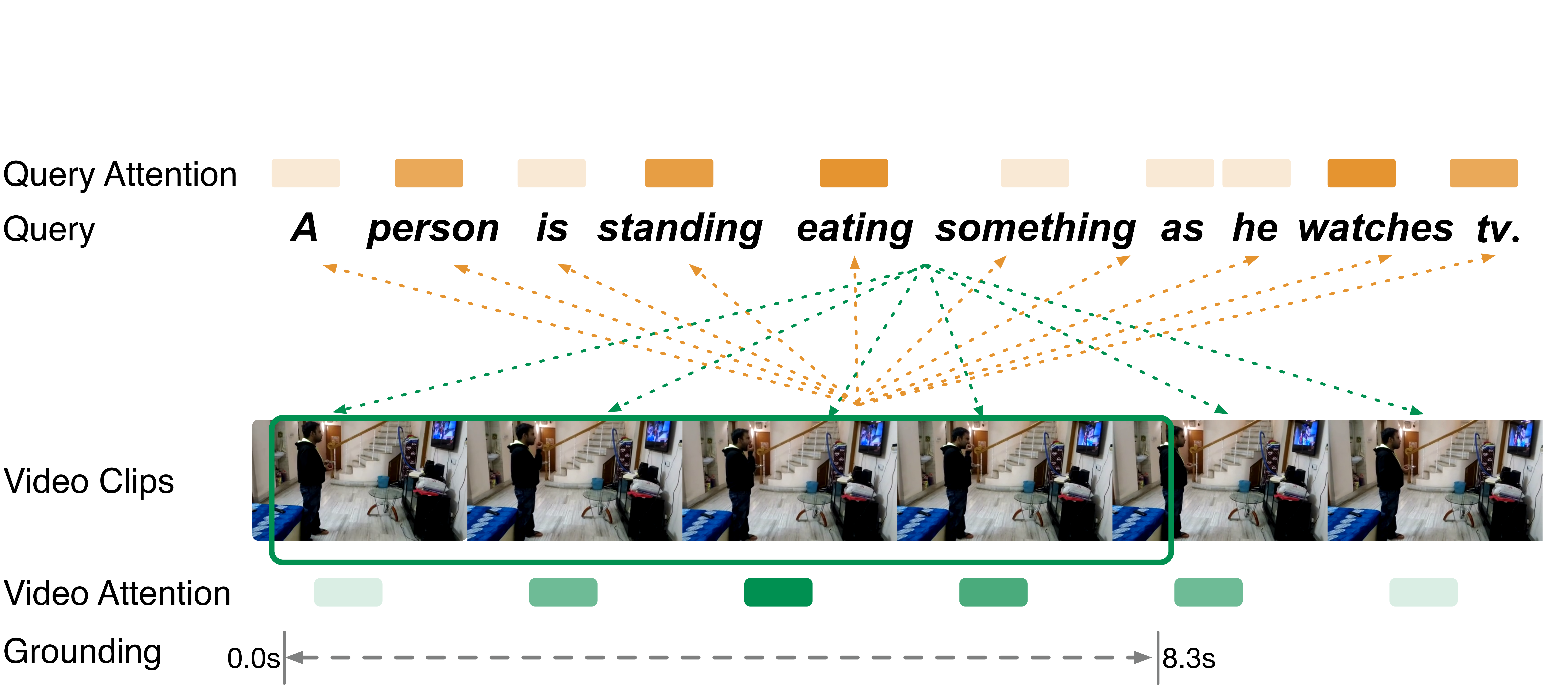}
    \caption{Illustration of Video Temporal Grounding with cross-modality attention. The particular video clip is selected with the guidance of the query, where the orange blocks show the attention value of the middle video clip for each word, and the green ones reflect the relations between the target word "eat" and each video clip.}
    \label{fig:sketchMap}
\end{figure}

    In this paper, we focus on Video Temporal Grounding (VTG) using a language query, which is first proposed by ~\cite{TALL:gao2017tall}. Formally, as illustrated in Figure ~\ref{fig:sketchMap}, given an untrimmed video and a language query, we need to localize the particular video clip, which contains the same semantic information corresponding with the query.

    Generally, localizing the video clip with a query contains three phases: input embeddings for video and query, multi-modality fusion, and localization. The multi-modality fusion methods are quite diverse and with no common baseline. We can briefly group these methods into attention-free~\cite{TALL:gao2017tall} and attention-based ~\cite{ABLR:yuan2019find, SCDM:yuan2019semantic,CBP:wang2020temporally}. Gao \etal. project video and text features into two global vectors and then use the fully-connected layer to get the fused features~\cite{TALL:gao2017tall}. It suffers from a lack of contextual information due to shallow fusion.

    As the attention mechanism is proved as an efficient method for feature extraction, Yuan \etal. use a multi-modality co-attention mechanism to generate attention attended with a global vector of another modality ~\cite{ABLR:yuan2019find}. Using a global vector as the representation of the query modality suffers the lack of local information when tackling with complex sentences. Mun \etal. propose a Sequential Query Attention Network to extract distinct phrases from the sentence, which contains more fine-grained semantics, and fuse the video with multiple phrases step by step. However, the fact is ignored that the video can also select and modulate the query sentence.

    While promising results have been achieved, there are still some challenges unsolved: 1. Although existing methods are getting higher performance, their modals become more and more complicated. 2. Considering the gaps between the vision and the language, aligning them both locally and globally need to be carefully conducted. 3. We notice the existing datasets contain inevitable annotation bias due to the ambiguity of action boundaries. The occurrence of action is a gradual process, including beginning, middle, and end. However, this continuous transition is not reflected in the action boundary which are just the start and end points.

    Usually, videos contain richer information than queries in the VTG task, so it is hard to capture the desired content with just one interaction. Inspired by the effectiveness of the coarse-to-fine strategy in searching for the answer to the question in long documents~\cite{choi2017coarse}, we conduct global and local interactions step by step to dig video information. In detail, we ﬁrst take a glimpse of the query, choose diverse semantic phrases, and then browse roughly the video sequence under the guide of phrases. When we ground the relative segments, details need to be compared carefully to ﬁnd the precise location. Specifically, we propose a coarse-to-fine encoder-decoder structure for VTG with Cross-Modality Attention (CMA), as depicted in Figure ~\ref{fig:structure}. In the encoder layer, video features are adjusted by diverse query guide vectors, which are extracted by our Semantic Phrase Extracting (SPE) Network, using multi-head self-attention. So the coarse-grained features are captured. In the decoder layer, the whole video sequence and all query words are refined alternately using self-attention and bi-attention mechanisms. Query features guide the video to focus on the main content. Meanwhile, video features highlight the decisive words in the query sentence, making the details in the two modalities align well.

    Methods to relate the fused modalities with the temporal location can be divided into two categories: anchor-based  ~\cite{SCDM:yuan2019semantic,CBP:wang2020temporally} and regression-based ~\cite {ABLR:yuan2019find,DRN:zeng2020dense,LGI:mun2020local}. Anchor-based methods calculate scores for pre-defined anchors and select the top recall anchors as the final output after using NMS ~\cite{NMS:neubeck2006efficient}. To avoid the redundant calculation, we adopt the regression-based method to predict the temporal location, and design a task-specific regularizer to minimize the effect of inherent annotation bias. Our model can tackle VTG efficiently through end-to-end training.

    The main contributions of our work are four-fold:

    \begin{itemize}
        \item We propose a simple pipeline for VTG using cross-modality attention with fewer parameters.
        \item We use a coarse-to-fine strategy to dig and align details in both videos and sentences.
        \item To our best knowledge, We first notice that datasets exist inevitable bias due to the ambiguity of action boundaries. To reduce the impact of annotation bias, we propose a new task-specific regression loss, which is robust for small location errors.
        \item We conduct extensive experiments to validate the effectiveness of our method and show that it can outperform the state of the arts on both Charades-STA and ActivityNet Captions datasets.
    \end{itemize}

\section{Related Work}
    \noindent In this section, we introduce several works in video temporal grounding, mainly focusing on different fusion methods and grounding strategies. Since the encoder-decoder structure adopted in the proposed method is widely used, we also list some related works from other vision-and-language tasks.

    \subsection{Video Temporal Grounding}
    \noindent Video temporal grounding is to localize the particular video clip guided by a language query. Early work~\cite{MCN:anne2017localizing} uses the “scan and localize” framework, where the specific clip is retrieved by scanning the whole video using sliding windows. Specifically, Hendricks \etal. build the Moment Context Network (MCN) to directly learn the correlation between the different video segments and the query~\cite{MCN:anne2017localizing}. This scheme is time-consuming and contains redundant calculations. In contrast, Gao \etal. extract two global vectors for the video and the query, then fuse them by a Fully-Connected (FC) layer~\cite{TALL:gao2017tall}. Since the FC layer is not enough to obtain the in-depth contextual information, attention mechanism are also incorporated in later  methods~\cite{ACRN:liu2018attentive,ABLR:yuan2019find,SCDM:yuan2019semantic}. 

    The concrete localization methods mainly belong to two categories, anchor-based~\cite{CBP:wang2020temporally,SCDM:yuan2019semantic}, and anchor-free~\cite{TALL:gao2017tall,DRN:zeng2020dense,TMLGA:rodriguez2020proposal}. Similar to two-stage object detection[], Yuan \etal. score for each pre-defined anchor and use NMS to select the top anchor as the prediction~\cite{SCDM:yuan2019semantic}. It’s noticeable that anchor-based methods neglect time dependencies across video clips and exist redundant calculations. To incorporate the temporal information between different clips, Yuan \etal. and Mun \etal. directly regresses the coordinates using MLP~\cite{ABLR:yuan2019find, LGI:mun2020local}. 
    Inspired by FCOS~\cite{FCOS:tian2019fcos}, Zeng \etal. design a dense regression network (DRN) with three heads to regress the distances from each frame to the start and end time~\cite{DRN:zeng2020dense}. 

\begin{figure*} [t]
    \begin{center}
        \includegraphics[width=\linewidth]{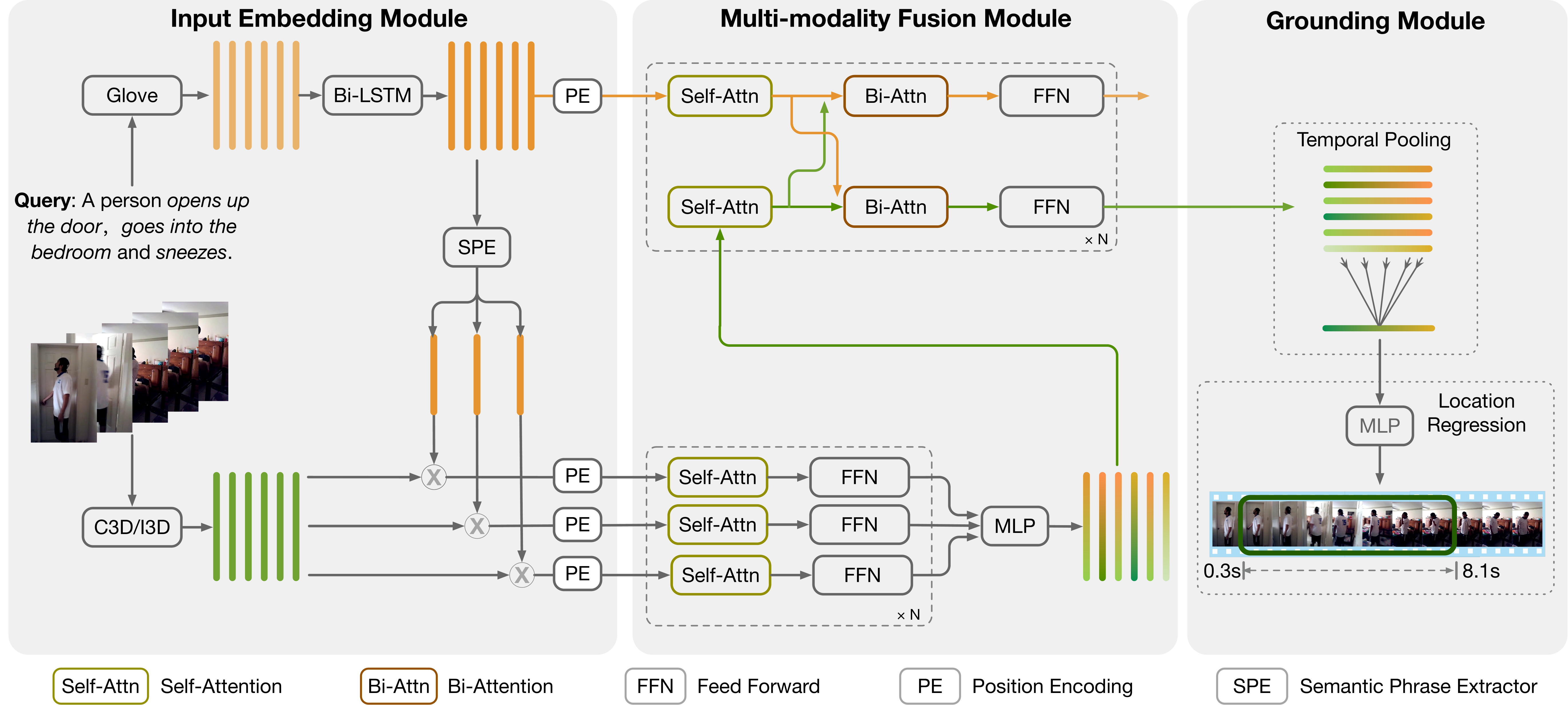}
    \end{center}
    \caption{
        The framework of our proposed method, which contains three main components. (1) Input embedding module encodes the original video and query. (2) Multi-modality fusion module fuses the information from two modalities. (3) Grounding module regresses the temporal location using the aligned features.
    }
    \label{fig:structure}
\end{figure*}

    \subsection{Encoder-Decoder Architecture}
    \noindent Videos and queries are both sequential, so the Seq2Seq models ~\cite{LSTM:hochreiter1997long,ATTENTION:vaswani2017attention} are suitable for dealing with the VTG task. Wang \etal. use Match-LSTM to aggregate contextual information by modeling the correlation between each frame and its neighbors~\cite{CBP:wang2020temporally}. However, LSTMs can not handle the long sequence dependency due to the information attenuation during forward propagation. Unlike previously described sequence-to-sequence models, Transformer uses the attention mechanism to calculate the whole sequence's weights for each element. Not only the long temporal dependency is solved, but also the processing speeds up because of parallel computation~\cite{dai2019transformer}.

    Promising performances are achieved with the Transformer in other multi-modality tasks, such as action recognition~\cite{ACTION:girdhar2019video}, video captioning~\cite{CAPTION:zhou2018end} and video question answering~\cite{VQABERT:yang2020bert}. Girdhar \etal. propose an action transformer to recognize and localize human actions in video clips, where contextual information from other humans and objects in the neighborhood frames is aggregated~\cite{ACTION:girdhar2019video}. Zhou \etal. design a dense video captioning model with Transformer~\cite{CAPTION:zhou2018end}. Specifically, the encoder encodes the video and detects the potential events; the decoder generates the descriptions using the output of the encoder. Yang \etal. jointly model the visual concepts and subtitles with a Transformer~\cite{VQABERT:yang2020bert}.

    Following the above works, we design an encoder-decoder transformer with cross-modality attention to fuse the multi-modal information for the VTG task.

\section{Methods}
    \noindent In this section, we introduce our main framework for video temporal grounding, as shown in Figure~\ref{fig:structure}. Our model consists of three primary components: the input embedding, the multi-modality fusion, and the grounding modules.

    \subsection{Problem Definition}
    Given an untrimmed video X and a language query Q, Video Temporal Grounding (VTG) is to localize the start time $\tau^s$ and end time $\tau^e$ of X corresponding to the Q.
    \begin{equation}
        (\tau^s,\tau^e) = f_\theta(X,Q).
    \end{equation}

    \subsection{Input Embedding}
        \subsubsection{Video Embedding}\label{sec:pe}
        \noindent Generally, static spatial information and dynamic temporal contents are both crucial for video understanding. Following the previous works~\cite{ABLR:yuan2019find,SCDM:yuan2019semantic,LGI:mun2020local}, we use 3D Conv. Network~\cite{C3D:tran2015learning,I3D:carreira2017quo}, denoted by $f_v(\cdot)$, to extract the spatio-temporal features from original video sequence $X$.

        Notice that the attention mechanism is permutation invariant~\cite{ATTENTION:vaswani2017attention}, so the Position Encoding (PE), denoted by $f_{PE}(\cdot)$ , is crucial to keep the sequential information. Two different PE methods are compared in this paper: constant sine/cosine position encoding and learnable position embedding. More details are showed in Section~\ref{sec:pe2}.
        The embedded video features $V^{in}=\{v_1^{in},v_2^{in},\dots,v_N^{in}\}$ can be represented as
        \begin{equation}
            V^{in} = ReLU(W_v f_v(X)) + f_{PE}([1,2,\dots,N]),
        \end{equation} where $W_v\in \mathbb{R}^{d\times d_v}, V^{in}\in \mathbb{R}^{d\times N}$. $d$ and $d_v$ are the dimensions of embedded video features and C3D features, respectively. $N$ represents the number of video clips.

        \subsubsection{Query Embedding}
        \noindent Each word in the query $Q=\{w_1,w_2,\dots,w_L\}$ is embedded to a 300D vector using pretrained Glove~\cite{GLOVE:pennington2014glove} word embedding. After getting the static word embedding, a bi-directional LSTM is conducted to encode contextual information $h_i$. We denote the embedded query sentence $Q^{in}\in \mathbb{R}^{d_q \times L}$ as the concatination of the forward LSTM and backward LSTM.
        The global sentence embedding $s_{global}$ consists of the concatination of the last $h_L^f$ and the first $h_1^b$ hidden states of forward and backward LSTMs.
        \begin{align}
            h^f_1,h^f_2,\dots,h^f_L &= LSTM^f(w_1,w_2,\dots,w_L), \\
            h^b_1,h^b_2,\dots,h^b_L &= LSTM^b(w_1,w_2,\dots,w_L), \\
            Q^{in} &= [h^f_i ; h^b_i] + f_{PE}([1,2,\dots,L]), \\
            s_{global} &= [h^f_L;h^b_1],
        \end{align}
        where $L$ is the length of the query sentence, and $[\cdot; \cdot]$ represents the concatenation operation. The dimension of the embedded query feature is $d_q$.

        \subsubsection{Semantic Phrase Extracting}
        \noindent A complex query sentence usually contains multiple semantic phrases, \eg, the sentence ``A person \textit{opens up the door}, \textit{goes into the bedroom} and \textit{sneezes}" consists of three main actions. Inspired by the Sequential Query Attention Network  (SQAN)~\cite{LGI:mun2020local}, we propose a Semantic Phrase Extracting Network (SPE), with attention following~\cite{LOSS:lin2017structured}. With SPE, we extract multiple query guide vectors, denoted by  $G = [g_{(1)};g_{(2)};\dots;g_{(k)}]$ from $Q^{in}$. These vectors coarsely lead the video to highlight different essential information.
        \begin{align}
            A_{spe} &= softmax(W_{s2}(tanh(W_{s1}Q^{in}))), \\
            G  &= Q^{in} A_{spe}^\top,
        \end{align}
        where $W_{s1}\in \mathbb{R}^{d_s\times d}$, $W_{s2}\in \mathbb{R}^{k\times d_s}$, $A_{spe}\in \mathbb{R}^{k\times L}$,$G\in \mathbb{R}^{d\times k}$, $k$ denotes the number of phrases.

        Note that the query guide vector is the same as $s_{global}$ when we choose a single phrase as the input of the encoder.

    \subsection{Multi-modality Fusion}
    \begin{figure} [t]
        \centering
        \includegraphics[width=\linewidth]{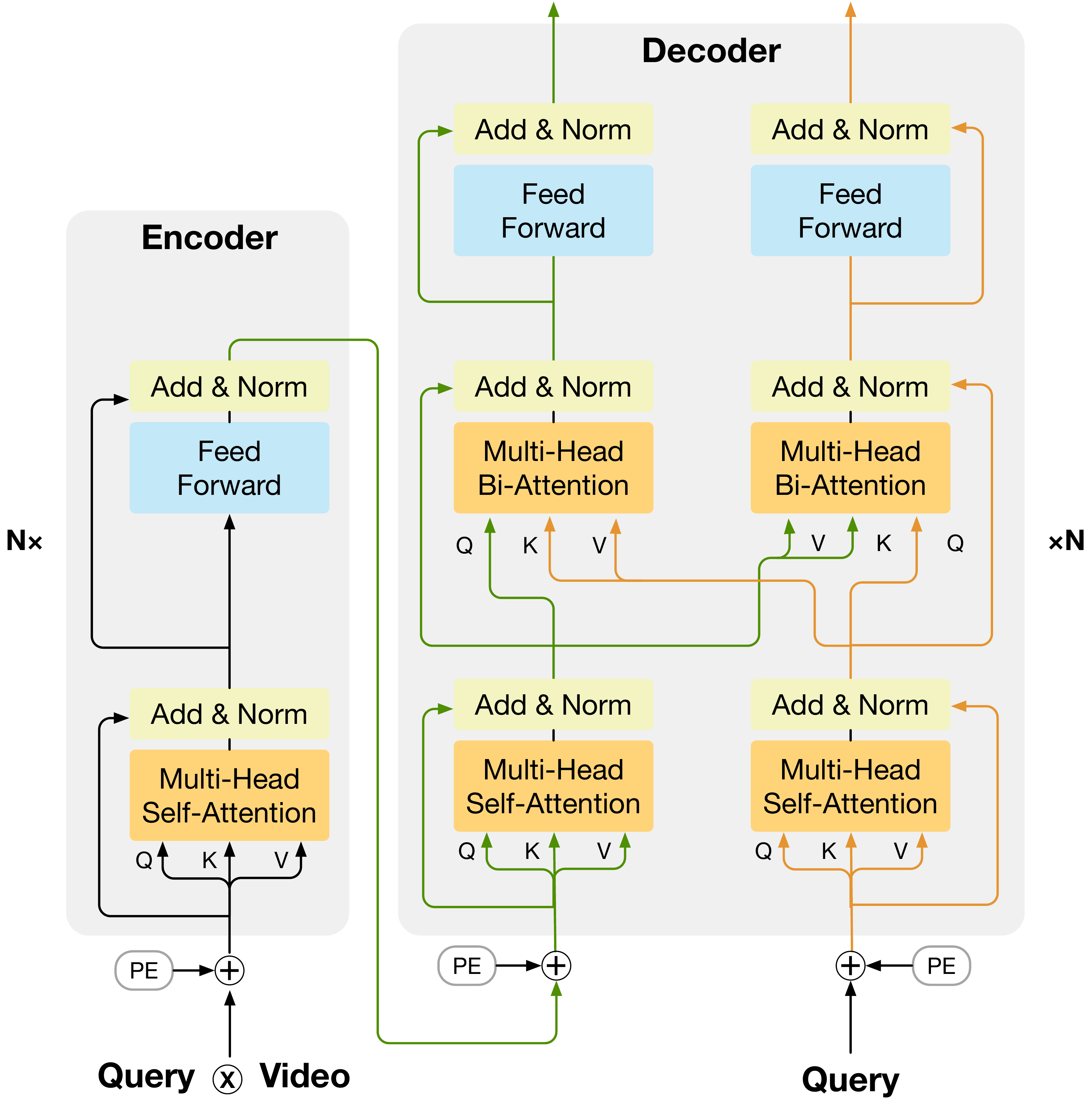}
        \caption{Details of multi-modality fusion module. (a) The encoder coarsely adjust the video features guided by query guide vectors; (b) The decoder consists of two cross-modality branches (left is the video branch and right is query branch), where two modalities are aligned alternatively. }
        \label{fig:transformer}
    \end{figure}
    \noindent The proposed cross-modality attention mechanism plays an essential role in aligning the local and global information, as depicted in Figure~\ref{fig:transformer}. Specifically, we adopt an encoder-decoder structure, where the video features $V^{in}$ are adjusted roughly by query guide vectors $g^{(i)}$ in the encoder, and local contents of video segments and query words are alternatively updated in the decoder.

        \subsubsection{Encoder}
        \noindent To correlate the semantic information of different modalities, we combine video features $V^{in}$ with each query guide vector $g^{(i)}$ to obtain the coarse video features $F^{in}$. Here we compare three different fusion methods, represented as $f_u(\cdot,\cdot)$, Hadamard Product, Concatenation, and Addition. More details refer to Section~\ref{sec:fusion}.
        \begin{equation}
            F^{in}_{(i)} = f_{u}(V^{in},g_{(i)}).
            \label{eq:fuse_early}
        \end{equation}

        The yielded representation $F^{in}_{(i)}$ is fed into a multi-head self-attention (SA), which is used to capture the interactions between two modalities in a coarse-grained manner. A MLP is used to aggregrate features adjusted by multiple query guide vectors.
        \begin{align}
            F^{enc}_{(i)} &= SA(F^{in}_{(i)}, F^{in}_{(i)}, F^{in}_{(i)}), \\
            F^{enc} &= MLP_{enc}([F^{enc}_{(1)},F^{enc}_{(2)},\dots,F^{enc}_{(n)}]).
        \end{align}

        For our single phrase model, Equation~\eqref{eq:fuse_early} is replaced by the following formula:
        \begin{equation}
            F^{in} = f_{u}(V^{in},s_{global}).
        \end{equation}

        \subsubsection{Decoder}
        \noindent Generally, not only the query can help the video to highlight the crucial information, but also the video can choose the keywords from the query. According to the above observation, we design two branches (query branch and video branch) cross-attention mechanism to modulate each modality alternately, and each branch consists of a self-attention and a bi-attention block.

        For the query branch, a self-attention takes all word embeddings $V^{in}$ into account, which extracts single modality information $F^{Q1}$ while keeps the original semitics. Then the words are attended by the enhanced video features $F^{V1}$ from another branch with a bi-attention component. The detail in the video branch is as same as the query branch. Two branches are parallel so multi modalities are alternatively aligned. The process of the decoder is given by
        \begin{align}
        F^{Q1} &= SA(Q^{in},Q^{in},Q^{in}), \\
        F^{V1} &= SA(F^{enc},F^{enc},F^{enc}), \\
        F^{Q2} &= BA(F^{Q1},F^{V1},F^{V1}), \\
        F^{V2} &= BA(F^{V1},F^{Q1},F^{Q1}),
        \end{align}
        where $F^{Q1},F^{Q2}\in \mathbb{R}^{d\times L}$, and $F^{V1},F^{V2}\in \mathbb{R}^{d\times N}$.

    \subsection{Grounding}
    \noindent After getting the fused features $F^{V2}$, there are two feasible ways to localize the clip, anchor-based and anchor-free. As mentioned before, the anchor-based method exists redundant calculation and breaks the temporal consistency, resulting in sub-optimal solutions. We choose the anchor-free grounding method, following ABLR~\cite{ABLR:yuan2019find}. Specifically, we design a temporal pooling module with attention mechanism to summarize the sequential information, and conduct a grounding block with a MLP to regress the boundaries $(\tau^s,\tau^e)$.
    \begin{align}
        B &= \tanh (W_{b} F^{V2}), \\
        a &= softmax(u_{ta}^\top B), \\
        \overline{f} &=\sum_{i=1}^N F^{V2}_i a_i, \\
        (\tau^s,\tau^e) &= MLP(\overline{f}),
    \end{align}
    where  $W_{b}\in \mathbb{R}^{d\times d}, u_{ta}\in \mathbb{R}^{d}, \overline{f}\in \mathbb{R}^{d}$.

    \subsection{Training}
    \noindent We denote the training set as $\{(X_i,Q_i,T_i,\tau^s_i,\tau^e_i)\}^K_{i=1}$, where $T_i$ represents the duration of the video $X_i$. And each description $Q_i$ matches one particular video clip in $X_i$, where the start and end points are $\tau^s_i$ and $\tau^e_i$ respectively.

    Our network is trained with three loss terms: (1) new location regression loss $L_{reg}$, (2) temporal attention loss $L_{ta}$, (3) semantic diversity loss, and the total loss is given by
    \begin{equation}
        L_{all} = L_{reg} + \lambda_{ta} L_{ta} + \lambda_{sd} L_{sd}.
    \end{equation}

    \subsubsection{Regression Loss}
    \noindent We notice that the action boundary can not be distinguished precisely. Different annotators have different standards, which results in the inevitable annotation bias in datasets. Since it is hard to define the exact event boundaries, our goal is to minimize the effect of the bias by introducing a new task-specific loss function.

    The quality of prediction depends on two main factors: the whole video duration and the ratio of ground truth interval to the duration. To handle with large-range video duration, Zeng \etal.  construct the feature pyramid~\cite{DRN:zeng2020dense}. In this paper, we normalize the start and end points $(\tau^s_i,\tau^e_i)$ into $[0,1]$ by dividing the video duration $T_i$, which is more efficient and less-calculation.
    \begin{equation}
        (\overline {t^s_i},\overline {t^e_i}) = (\tau^s_i/T_i,\tau^e_i/T_i).
    \end{equation}

    Previous works~\cite{ABLR:yuan2019find,DRN:zeng2020dense,LGI:mun2020local} directly adopt the Smooth L1 Loss.

    \begin{equation}
        Smooth_{L1} = \left\{
                        \begin{aligned}
                        0.5 x^2, && |x|<1 \\
                        |x| - 0.5, &&  |x|>1.\\
                        \end{aligned}
                        \right.
    \end{equation}

\begin{figure}[t]
        \centering
        \includegraphics[width=1\linewidth]{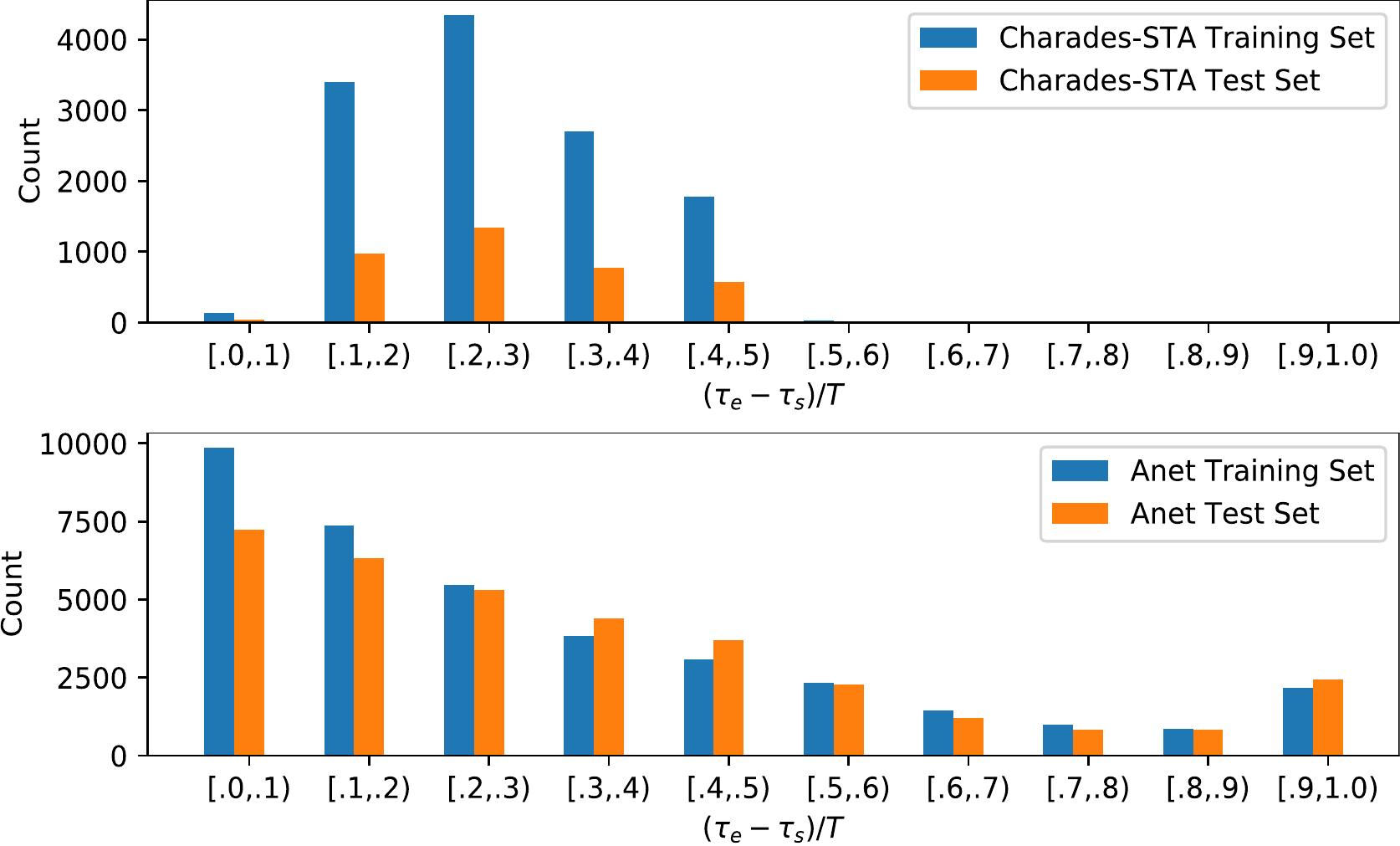}
        \caption{The distribution of ground-truth annotations in Charades-STA and ActivityNet Captions.}
        \label{fig:statistic}
    \end{figure}

    However, the normalized time points $(\overline {t^s_i},\overline {t^e_i})$ are less than 1, which actually becomes $L_2$ Loss in the region. Though $L_2$ loss can converge faster than $L_1$, $L_1$ loss is robust for outliers than $L_2$ loss~\cite{FASTRCNN:girshick2015fast}. To bare with outliers and avoid the gradient vanishing in the meanwhile, we introduce a dataset-specific threshold $\beta \in (0,1)$ between $L_1$ and $L_2$. In order to choose the best value, we count the ratio of the ground-truth interval to the whole video duration in Charades-STA and ActivityNet Captions datasets, as shown in Figure~\ref{fig:statistic}. The results show that the average ratios are 0.27 and 0.33. To control the accuracy of prediction, we introduce the coefficient $\alpha$ to weight the task-specific regression loss, $f(x)$.

    Concrete deﬁnitions are as follows:
    \begin{equation}
        f(x) = \left\{
                \begin{aligned}
                \alpha x^2, && |x|<\beta \\
                2\alpha \beta |x| - \alpha \beta^2, &&  |x|>\beta.\\
                \end{aligned}
                \right.
    \end{equation}

    \subsubsection{Temporal Attention Loss}
    \noindent In the grounding module, we first conduct a temporal attention pooling to summarize the fused features. Generally, the components in ground truth play an essential role in localizing, so the attention weights $a_{ta}$ related to the target are expected to be higher than others. Since there is no explicit ground truth of attention, we adopt a temporal attention loss following ~\cite{ABLR:yuan2019find}, which is given by

    \begin{gather}
        L_{ta} = -\frac{\sum^N_{i=1} \mathbbm{1}(a_i) \log(a_i)}{\sum_{i=1}^N \mathbbm{1}(a_i)},
    \end{gather} where $\mathbbm{1}(a_i)=1$ if the $i$-th feature is in the ground truth of boundaries and 0 otherwise.

    \subsubsection{Semantic Diversity Loss}
    \noindent To keep the semantic diversity between different phrases, we conduct a regularization term on the attention matrix $A_{spe}$ obtained by SPE, following~\cite{LOSS:lin2017structured}.
    \begin{equation}
        L_{sd} = ||A_{spe}A_{spe}^\top - I||_F^2,
    \end{equation}
    where $A_{spe}\in \mathbb{R}^{k\times L}$ , and $||\cdot||_F$ stands for the Frobenius norm of a matrix.

\section{Experiments}
    \noindent We demonstrate the feasibility of our CMA model on two public datasets: Charades-STA~\cite{TALL:gao2017tall}, and ActivityNet Captions ~\cite{ACTIVITYNET:krishna2017dense}. Then we evaluate the improvement of performance in different modules.

    \subsection{Datasets}
        \subsubsection{Charades-STA}
        \noindent consists of around 10,000 videos, and each video contains activity annotation and video-level description but without clip-level annotation. ~\cite{TALL:gao2017tall} design a semi-automatic method to generate the description annotation with a start and end time. In total, the Charades-STA dataset contains 6,672 videos, 16,128 clip-sentence pairs, and 157 activity categories. There are 12,408 / 3,720 pairs in the training / test set. On average, the duration of videos is around 30s, and the words of each query are about 10.

        \subsubsection{ActivityNet Captions}
        \noindent contains 20,000 videos with 100,000 queries, covers 200 activity classes.  It is split into 37421, 17505, 17031 pairs for training, validation, and test. On average, the duration of each video is 2 minutes and the length of each query is 13.48 words. Since the annotations of the test set are non-public, we merge the two validation sets “val\_1” and “val\_2” as our test split as previous works do~\cite{LGI:mun2020local,CBP:wang2020temporally}.

\begin{table}

    \caption{
    Experimental results on the Charades-STA dataset. $^+$ denotes the full model with multiple guide vectors.
    }
    \label{tab:charades}
    \centering
    \begin{tabular}{lccccc}
		    \toprule
            Method     &  R1@0.3    & R1@0.5  & R1@0.7  & mIoU \\
            \midrule
            CTRL & - & 21.42 & 7.15 & - \\
            SMRL & - & 24.36 & 9.01 & -\\
            RWM & - & 36.70 & - & - & - \\
            ExCL & 65.10 & 44.10 & 22.60 & -\\
            CBP & - & 36.80 & 18.87 & -\\
            SCDM & - & 54.44 & 33.43 & -\\
            TMLGA & 67.53 & 52.02 & 33.74 & -\\
            DRN & - & 53.09 & 31.75 & -\\
            \midrule
            LGI & 71.02  & 57.34 & 33.25 & 49.52\\
            Ours   &  \bf{71.38}   & \bf{57.56} & \bf{35.18}  & \bf{50.01} \\
            \midrule
            LGI$^+$ & \bf{72.96} & 59.46 & 35.48 & \bf{51.38} \\

            Ours$^+$   &  72.20   & \bf{59.81} & \bf{37.72}  & 51.04 \\
            \bottomrule
    \end{tabular}
\end{table}

    \subsection{Metrics}
    \noindent We choose similar metrics ``R@$n$, IoU@$m$" and ``mIoU" from ~\cite{TALL:gao2017tall} to evaluate the performance of our model. The Intersection over Union (IoU) between the prediction and ground truth is calculated for each query. ``R@$n$, IoU@$m$" means the percentage of at least one of the top-$n$ retrievals larger than the threshold $m$. We choose $n=\{1\},m=\{0.3,0.5,0.7\}$ and abbreviate the symbol as ``R$n$@$m$". ``mIoU" denotes the average IoU for all queries.

    \subsection{Implementation details}
        \subsubsection{Video Feature}
        \noindent We uniformly sample $N=128$ clips from each video and each clip contains 16 frames. Then we adopt the C3D~\cite{C3D:tran2015learning} network as the feature extractor of ActivityNet Captions. We also extract the I3D~\cite{I3D:carreira2017quo} features for Charades-STA.

        \subsubsection{Language Feature}
        \noindent We first tokenize and lowercase the query, and use Glove~\cite{GLOVE:pennington2014glove} to get the static word embedding with 300D. Then a two-layer bi-directional LSTM is used to get the contextual word embedding, whose hidden layer dimension is 256. The maximum query words for Charades-STA and ActivityNet Captions are 10 and 25, respectively. Vocabulary sizes are 1,140 and 11,125.

        \subsubsection{Training Settings}
        \noindent Empirically, We set mini-batch size as 100, optimizer as Adam and initial learning rate as $1\times 10^{-3}$. The parameters $\alpha,\beta$ in the task-specific regression loss are set $10,0.1$ for Charades-STA; $2,0.4$ for ActivityNet Captions, respectively. We use two-layer encoder-decoder with 4 heads for two datasets. The trade-off parameters $\lambda_{ta}$ and $\lambda_{sd}$ are set to 1.

\subsection{Results}

     \begin{table}

    \caption{Experimental results on the ActivityNet Captions dataset. $^+$ denotes the full model with multiple guide vectors.}
    \label{tab:activitynet}
    \centering
    \begin{tabular}{lcccc}
        \toprule
        Method     &  R1@0.3    & R1@0.5  & R1@0.7  & mIoU \\
        \midrule
        MCN & 21.37 & 9.58 & - & 15.83  \\
        CTRL & - & 21.42 & 7.15 & - \\
        ACRN & 31.29 & 16.17 & - & 24.16\\
        RWM & - & 36.90 & - & - \\
        ABLR & 55.67 & 36.79 & - & 36.99\\
        CBP & 54.30 & 35.76 & 17.80 & - \\
        SCDM & 54.80 & 36.75 & 19.86 & - \\
        TMLGA & - & - & - & 37.78\\
        \midrule
        LGI & 57.79 & 40.97 & 23.16 & 40.74 \\
        Ours   &  \bf{58.67}  & \bf{ 41.43} & \bf{23.59}  & \bf{41.23}   \\

        \midrule
        LGI$^+$ & 58.52 & 41.51 & 23.07 & 41.13 \\
        Ours$^+$ & \bf{59.10} & \bf{41.93} & \bf{24.23} & \bf{41.96} \\
        \bottomrule
    \end{tabular}
\end{table}

   \begin{figure}
        \centering
            \vspace{0.8cm}

        \includegraphics[width=\linewidth]{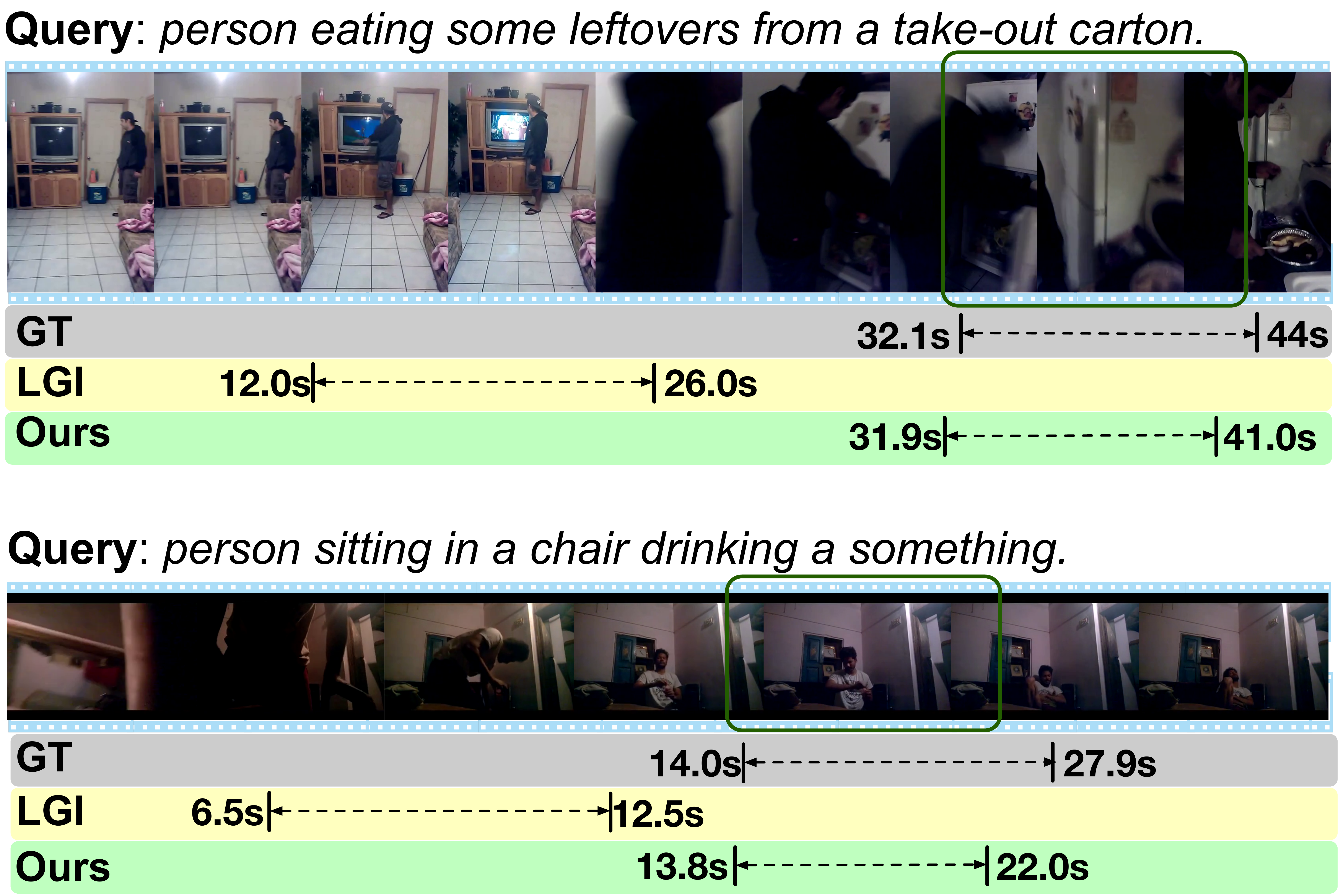}
        \caption{Qualitative results of LGI and our methods on Charades-STA.}
        \label{fig:qualitaive}
    \end{figure}

    We compare our proposed method with the following: MCN~\cite{MCN:anne2017localizing}, CTRL~\cite{TALL:gao2017tall}, SMRL~\cite{SMRL:wang2019language}, RWM~\cite{RWM:he2019read}, ExCL~\cite{EXCL:ghosh2019excl}, CBP~\cite{CBP:wang2020temporally}, SCDM~\cite{SCDM:yuan2019semantic}, TMLGA~\cite{TMLGA:rodriguez2020proposal}, DRN~\cite{DRN:zeng2020dense}, and LGI~\cite{LGI:mun2020local}.

    \subsubsection{Charades-STA}
    \noindent Table~\ref{tab:charades} summarizes performances on Charades-STA. Notice that our simple model (without SPE) makes the relative improvement over LGI~\cite{LGI:mun2020local} by $0.36\%$ (R1@0.3), $0.22\%$ (R1@0.5), and $1.93\%$ (R1@0.7). Our full model also outperforms LGI$^+$ by $0.35\%$ (R1@0.5) and $2.24\%$ (R1@0.7).

    \subsubsection{ActivityNet Captions}
    \noindent The numerical results are shown in Table~\ref{tab:activitynet}. As can be seen, our simple and full models outperform the counterparts in LGI respectively. Note that our model has only less than half of the number of the parameters, which demonstrates the efficiency of our CMA.
    \subsubsection{Qualitative Results}
\begin{figure*}
        \centering
        \includegraphics[width=\linewidth]{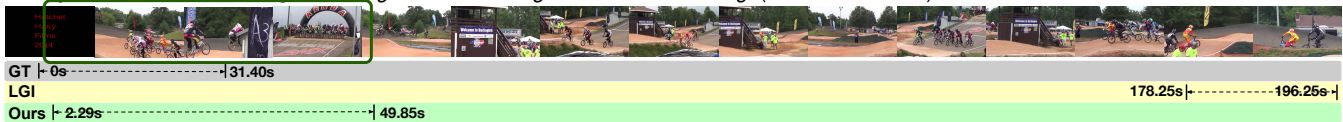}
        \caption{Qualitative results of LGI and our methods on ActivityNet Captions.}
        \label{fig:qualitaive2}
\end{figure*}

    We provide some qualitative examples to validate the effectiveness of the CMA module. As shown in Figure~\ref{fig:qualitaive} and Figure~\ref{fig:qualitaive2}, our method can localize the approximate boundaries even for complicated situations like long-duration videos and complex sentences, while LGI fails.

    \subsection{Ablation Studies}

          \begin{table}
        \caption{Performance comparison for different structures on the Charades-STA dataset.}
            \begin{center}
                \setlength{\tabcolsep}{29 pt}
                \begin{tabular}{lc}
                    \toprule
                    Method & R1@0.5  \\
                    \midrule
                    Encoder-only & 51.16   \\
                    Decoder-only & 55.81  \\
                    Full model & \bf{57.56} \\

                    \bottomrule
                \end{tabular}
            \end{center}
            \label{tab:transformer}
        \end{table}

        \begin{table}
        \caption{Performance comparison for different position encoding methods on the Charades-STA dataset.}
            \begin{center}
                \setlength{\tabcolsep}{19 pt}
                \begin{tabular}{lc}
                    \toprule
                    Method & R1@0.5  \\
                    \midrule
                    Ours-w/o PE & 53.63 \\
                    Ours-Embedding matrix & 56.29   \\
                    Ours-Sin/cos PE & \bf{57.56} \\
                    \bottomrule
                \end{tabular}
            \end{center}
            \label{tab:pe}
        \end{table}

    \noindent Our CMA model consists of multiple components, including bi-modality encoder-decoder transformer and location regression. To evaluate the contribution of each module to the final performance, we conduct three primary ablation studies on Charades-STA dataset. We employ the simple model  (without SPE) to demonstrate the efficiency of our cross-modality fusion method.

        \subsubsection{Encoder-Decoder}
        \noindent We ﬁrst investigate the contribution of the coarse-to-fine strategy. In this experiment, we train three variants of our model: (1) Encoder-only: we only use encoder layers to coarsely interact with two modalities; (2) Decoder-only: we just align the local information with decoder layers; (3) Encoder-decoder: our full model perform multi-modality fusion based on encoder-decoder structure.

        Table~\ref{tab:transformer} summarizes the results where we observe that our full model outperforms the other variants. It is due to two reasons. First, encoder-only use the global sentence information, which is not enough to localize the precise time interval due to the lack of local details. Second, decoder-only is easy to stuck with sub-optical parameters without the guide of global information. This shows both global and regional knowledge are both crucial.

        \subsubsection{Position Encoding}\label{sec:pe2}
        \noindent To evaluate the effectiveness of the temporal position embedding, we conduct three experiments: (1)w/o PE: features are fed into the network without adding PE. (2) use sine and cosine PE following~\cite{ATTENTION:vaswani2017attention}, see Equal~\eqref{eq:PE}. (3) learn two position embedding matrixes $W_{pos}^V\in \mathbb{R}^{d\times N}, W^Q_{pos}\in \mathbb{R}^{d\times L}$. The result in Table~\ref{tab:pe} shows that PE is essential for precise localization because the transformer is non-sensitive to the permutation. Another point is that two different PE methods reach similar performance, so we choose the former because it’s efficient with less parameters.
        \begin{gather}
            PE_{(pos,2i)} = sin(\frac{pos}{10000^{2i/model}}), \\
            PE_{(pos,2i+1)} = cos(\frac{pos}{10000^{2i/model}}). \label{eq:PE}
        \end{gather}

        \subsubsection{Fusion Methods} \label{sec:fusion}
        Here we compare three combination methods ($f_u(\cdot,\cdot)$): Hadamard Product, Linear Addition, and Concatenation. The results in Table~\ref{tab:fusion} show that Hadamard Product achieves the best performance.

        \subsubsection{Loss Design}
       According to the distribution in Figure~\ref{fig:statistic}, we choose the threshold $\beta$ from $\{0.1,0.2,1\}$. We find that the performance is improved when the threshold is close to the average of the ratio. In addition, properly increasing the value of the coefficient $\alpha$ can also improve performance. We compare different loss settings and find ours achieves the best performance as listed in Table~\ref{tab:reg_loss}.

        \begin{table}
        \caption{Performance comparison for different fusion methods on the Charades-STA dataset.}
            \begin{center}
                \setlength{\tabcolsep}{23 pt}
                \begin{tabular}{lc}
                    \toprule
                    Method & R1@0.5\\
                    \midrule
                    Ours-Add & 53.42   \\
                    Ours-Concat & 56.45   \\
                    Ours-HadamardProduct & \bf{57.56} \\
                    \bottomrule
                \end{tabular}
            \end{center}
            \label{tab:fusion}
        \end{table}

        \begin{table}
        \caption{Performance comparison for different loss functions on the Charades-STA dataset.}
            \begin{center}
                \setlength{\tabcolsep}{15 pt}
                \begin{tabular}{cccc}
                    \toprule
                    Reg. Loss & $\alpha$ & $\beta$ & R1@0.5\\
                    \midrule
                    $Smooth_{L1}$ & 0.5 & 1 & 56.45 \\
                    Ours & 2 & 0.1 & 56.59 \\
                    Ours & 2 & 0.2 & 57.07 \\
                    Ours & 10 & 0.1 & \bf{57.56} \\
                    \bottomrule
                \end{tabular}
            \end{center}
            \label{tab:reg_loss}
        \end{table}

\section{Conclusion}
    \noindent In this paper, we propose a simple but powerful model for video temporal grounding with two-branch cross-modality attention. In this scheme, multi modalities are alternatively modulated, so the local and global contents are aligned well. In the meanwhile, we design a new regression loss that alleviate the effect of the annotation bias. The promising performances achieved on Charades-STA and ActivityNet Captions demonstrate the effectiveness of our method.

\newpage
\bibstyle{aaai}
\bibliography{bib}

\end{document}